\pdfoutput=1

\documentclass[11pt,letterpaper]{article}
\usepackage{authblk}
\usepackage[letterpaper]{geometry}
\usepackage{acl2012}
\usepackage{times}
\usepackage{latexsym}
\usepackage{amsmath}
\usepackage{amsfonts}
\usepackage{multirow}
\usepackage{url}

\usepackage{graphicx}
\usepackage{dcolumn}
\usepackage{bm}

\usepackage{capt-of}
\usepackage[page, title]{appendix}
\usepackage{rotating}
\usepackage{placeins}
\usepackage{color}

\graphicspath{{figures/}}

\newcommand{\specialcell}[2][l]{%
  \begin{tabular}[#1]{@{}l@{}}#2\end{tabular}}

\makeatletter
\newcommand{\@BIBLABEL}{\@emptybiblabel}
\newcommand{\@emptybiblabel}[1]{}
\makeatother
\usepackage[hidelinks]{hyperref}

\title{Anchored Correlation Explanation: \\Topic Modeling with Minimal Domain Knowledge}

\author[1,2]{\textbf{Ryan J. Gallagher}}
\author[1]{\textbf{Kyle Reing}}
\author[1]{\textbf{David Kale}}
\author[1]{\textbf{Greg Ver Steeg}}
\affil[1]{Information Sciences Institute, University of Southern California}
\affil[2]{Vermont Complex Systems Center, Computational Story Lab, University of Vermont\protect\\ {\tt ryan.gallagher@uvm.edu} \protect\\ {\tt\{\tt reing,kale,gregv\}@isi.edu}}

\date{}

\begin{document}

\maketitle

\begin{abstract}
While generative models such as Latent Dirichlet Allocation (LDA) have proven fruitful in topic modeling, they often require detailed assumptions and careful specification of hyperparameters. Such model complexity issues only compound when trying to generalize generative models to incorporate human input. We introduce Correlation Explanation (CorEx), an alternative approach to topic modeling that does not assume an underlying generative model, and instead learns maximally informative topics through an information-theoretic framework. This framework naturally generalizes to hierarchical and semi-supervised extensions with no additional modeling assumptions. In particular, word-level domain knowledge can be flexibly incorporated within CorEx through anchor words, allowing topic separability and representation to be promoted with minimal human intervention. Across a variety of datasets, metrics, and experiments, we demonstrate that CorEx produces topics that are comparable in quality to those produced by unsupervised and semi-supervised variants of LDA.
\end{abstract}


\section{Introduction}

The majority of topic modeling approaches utilize probabilistic generative models, models which specify mechanisms for how documents are written in order to infer latent topics. These mechanisms may be explicitly stated, as in Latent Dirichlet Allocation (LDA) \cite{blei2003latent}, or implicitly stated, as with matrix factorization techniques \cite{hofmann1999probabilistic,ding2008equivalence,buntine2006discrete}. The core generative mechanisms of LDA, in particular, have inspired numerous generalizations that account for additional information, such as the authorship \cite{rosen2004author}, document labels \cite{mcauliffe2008supervised}, or hierarchical structure \cite{griffiths2004hierarchical}.

However, these generalizations come at the cost of increasingly elaborate and unwieldy generative assumptions. While these assumptions allow topic inference to be tractable in the face of additional metadata, they progressively constrain topics to a narrower view of what a topic can be. Such assumptions are undesirable in contexts where one wishes to minimize model complexity and learn topics without preexisting notions of how those topics originated.

For these reasons, we propose topic modeling by way of Correlation Explanation (CorEx),\footnote{Open source, documented code for the CorEx topic model available at \url{https://github.com/gregversteeg/corex_topic}.
} 
an information-theoretic approach to learning latent topics over documents. Unlike LDA, CorEx does not assume a particular data generating model, and instead searches for topics that are ``maximally informative" about a set of documents. By learning informative topics rather than generated topics, we avoid specifying the structure and nature of topics ahead of time.

In addition, the lightweight framework underlying CorEx is versatile and naturally extends to hierarchical and semi-supervised variants with no additional modeling assumptions. More specifically, we may flexibly incorporate word-level domain knowledge within the CorEx topic model. Topic models are often susceptible to portraying only dominant themes of documents. Injecting a topic model, such as CorEx, with domain knowledge can help guide it towards otherwise underrepresented topics that are of importance to the user. By incorporating relevant domain words, we might encourage our topic model to recognize a rare disease that would otherwise be missed in clinical health notes, focus more attention to topics from news articles that can guide relief workers in distributing aid more effectively, or disambiguate aspects of a complex social issue.

Our contributions are as follows: first, we frame CorEx as a topic model and derive an efficient alteration to the CorEx algorithm to exploit sparse data, such as word counts in documents, for dramatic speedups. Second, we show how domain knowledge can be naturally integrated into CorEx through ``anchor words" and the information bottleneck. Third, we demonstrate that CorEx and anchored CorEx produce topics of comparable quality to unsupervised and semi-supervised variants of LDA over several datasets and metrics. Finally, we carefully detail several anchoring strategies that highlight the versatility of anchored CorEx on a variety of tasks.


\section{Methods}

\subsection{CorEx: Correlation Explanation}

Here we review the fundamentals of Correlation Explanation (CorEx), and adopt the notation used by Ver Steeg and Galstyan in their original presentation of the model \shortcite{ver2014discovering}. Let $X$ be a discrete random variable that takes on a finite number of values, indicated with lowercase, $x$. Furthermore, if we have $n$ such random variables, let $X_G$ denote a sub-collection of them, where $G \subseteq \{1, \ldots, n\}$.
The probability of observing $X_G=x_G$ is written as $p(X_G=x_G)$, which is typically abbreviated to $p(x_G)$. 
The entropy of $X$ is written as $H(X)$ and the mutual information of two random variables $X_1$ and $X_2$ is given by $I(X_1 : X_2) = H(X_1) + H(X_2) - H(X_1, X_2)$.

The total correlation, or multivariate mutual information, of a group of random variables $X_G$ is expressed as
\begin{align}
TC(X_G) &= \sum_{i \in G} H(X_i) - H(X_G) \label{total correlation}\\
	&= D_{KL} \left( p(x_G) || \prod_{i \in G} p(x_i) \right) \label{kl divergence}.  
\end{align}
We see that Eq.~\ref{total correlation} does not quantify ``correlation'' in the modern sense of the word, and so it can be helpful to conceptualize total correlation as a measure of total dependence. Indeed, Eq.~\ref{kl divergence} shows that total correlation can be expressed using the Kullback-Leibler Divergence and, therefore, it is zero if and only if the joint distribution of $X_G$ factorizes, or, in other words, there is no dependence between the random variables.

The total correlation can be written when conditioning on another random variable $Y$, $TC(X_G \mid Y) = \sum_{i \in G} H(X_i \mid Y) - H(X_G \mid Y)$. So, we can consider the reduction in the total correlation when conditioning on $Y$.
\begin{align}
TC(X_G ; Y) &= TC(X_G) - TC(X_G \mid Y) \label{lower bound} \\
	&= \sum_{i \in G} I(X_i : Y) - I(X_G : Y) \label{lower bound mutual information}
\end{align}
The quantity expressed in Eq.~\ref{lower bound} acts as a lower bound of $TC(X_G)$ \cite{ver2015maximally}, as readily verified by noting that $TC(X_G)$ and $TC(X_G|Y)$ are always non-negative. Also note, the joint distribution of $X_G$ factorizes conditional on $Y$ if and only if $TC(X_G \mid Y) = 0$. If this is the case, then $TC(X_G; Y)$ is maximized, and $Y$ explains all of the dependencies in $X_G$. 

In the context of topic modeling, $X_G$ represents a group of word types and $Y$ represents a topic to be learned. Since we are always interested in grouping multiple sets of words into multiple topics, we will denote the binary latent topics as $Y_1, \ldots Y_m$ and their corresponding groups of word types as $X_{G_j}$ for $j = 1, \ldots, m$ respectively. The CorEx topic model seeks to maximally explain the dependencies of words in documents through latent topics by maximizing $TC(X;Y_1,\ldots,Y_m)$. To do this, we maximize the following lower bound on this expression:
\begin{equation}
\max_{G_j, p(y_j \mid x_{G_j})}  \sum_{j = 1}^m TC(X_{G_j} ; Y_j). \label{corex objective}
\end{equation}
As we describe in the following section, this objective can be efficiently approximated, despite the search occurring over an exponentially large probability space \cite{ver2014discovering}.

Since each topic explains a certain portion of the overall total correlation, we may choose the number of topics by observing diminishing returns to the objective. Furthermore, since the CorEx implementation depends on a random initialization (as described shortly), one may restart the CorEx topic model several times and choose the one that explains the most total correlation.

The latent factors, $Y_j$, are optimized to be informative about dependencies in the data and do not require generative modeling assumptions. Note that the discovered factors, $Y$, can be used as inputs to construct new latent factors, $Z$, and so on leading to a hierarchy of topics. Although this extension is quite natural, we focus our analysis on the first level of topic representations for easier interpretation and evaluation.

\subsection{CorEx Implementation}

We summarize the implementation of CorEx as presented by Ver Steeg and Galstyan \shortcite{ver2014discovering} in preparation for innovations introduced in the subsequent sections. The numerical optimization for CorEx begins with a random initialization of parameters and then proceeds via an iterative update scheme similar to EM. For computational tractability, we subject the optimization in Eq.~\ref{corex objective} to the constraint that the groups, $G_j$, do not overlap, i.e. we enforce single-membership of words within topics. The optimization entails a combinatorial search over groups, so instead we look for a form that is more amenable to smooth optimization. We rewrite the objective using the alternate form in Eq.~\ref{lower bound mutual information} while introducing indicator variables $\alpha_{i,j}$ which are equal to 1 if and only if word $X_i$ appears in topic $Y_j$ (i.e. $i \in G_j$).
\begin{align}
&\max_{\alpha_{i,j}, p(y_j \mid x)} & &\sum_{j = 1}^m \left(\sum_{i =1}^n \alpha_{i,j} I(X_i : Y_j) - I(X: Y_j)\right) \nonumber \\
&\mbox{s.t.} &  &\alpha_{i,j} = \mathbb{I}[j= \arg \max_{\bar j} I(X_i:Y_{\bar j})]. \label{almost info bottleneck}
\end{align}
Note that the constraint on non-overlapping groups now becomes a constraint on $\alpha$. 
To make the optimization smooth we should relax the constraint so that $\alpha_{i,j} \in [0,1]$. To do so, we replace the second line with a softmax function. The update for $\alpha$ at iteration $t$ becomes,
$$
\alpha_{i,j}^{t} = \exp \left(\lambda^t (I(X_i:Y_j) - \max_{\bar j} I(X_i:Y_{\bar j}))\right).
$$
Now $\alpha \in [0,1]$ and the parameter $\lambda$ controls the sharpness of the softmax function. Early in the optimization we use a small value of $\lambda$, then increase it later in the optimization to enforce a hard constraint.
The objective in Eq.~\ref{almost info bottleneck} only lower bounds total correlation in the hard max limit. The constraint on $\alpha$ forces competition among latent factors to explain certain words, while setting $\lambda=0$ results in all factors learning the same thing. 
Holding $\alpha$ fixed, taking the derivative of the objective (with respect to the variables $p(y_j|x)$, and setting it equal to zero leads to a fixed point equation. We use this fixed point to define update equations at iteration $t$. 
\begin{align}
p_t(y_j) &= \sum_{\bar x} p_t(y_j|\bar x) p(\bar x) \label{marginals} \\
p_t(x_i|y_j) &=  \sum_{\bar x} p_t(y_j|\bar x) p(\bar x) \mathbb{I}[\bar x_i = x_i] / p_t(y_j) \nonumber \\
\log p_{t+1}(y_j|x^\ell) &= \label{self consistent eqn}\\
\log p_t(y_j) + &\sum_{i = 1}^n  \alpha^t_{i,j} \log \frac{p_t(x_i^\ell \mid y_j)}{p(x_i^\ell)} - \log \mathcal Z_j(x^\ell)  \nonumber
\end{align}
The first two lines just define the marginals in terms of the optimization parameter, $p_t(y_j|x)$. We take $p(x)$ to be the empirical distribution defined by some observed samples, $x^\ell, \ell=1,\ldots,N$. 
The third line updates $p_t(y_j|x^\ell)$, the probabilistic labels for each latent factor, $Y_j$, for a given sample, $x^\ell$. Note that an easily calculated constant, $\mathcal Z_j(x^\ell)$, appears to ensure the normalization of $p_t(y_j|x^\ell)$ for each sample.
We iterate through these updates until convergence. 

After convergence, we use the mutual information terms $I(X_i:Y_j)$ to rank which words are most informative for each factor. The objective is a sum of terms for each latent factor and this allows us to rank the contribution of each factor toward our lower bound on the total correlation. 
The expected log of the normalization constant, often called the free energy, $\mathbb E [\log \mathcal Z_j(x)$], plays an important role since its expectation provides a free estimate of the $i$-th term in the objective~\cite{ver2015maximally}, as can be seen by taking the expectation of Eq.~\ref{self consistent eqn} at convergence and comparing it to Eq.~\ref{almost info bottleneck}. Because our sample estimate of the objective is just the mean of contributions from individual sample points, $x^\ell$, we refer to $\log \mathcal Z_j(x^\ell)$ as the pointwise total correlation explained by factor $j$ for sample $\ell$. Pointwise TC can be used to localize which samples are particularly informative about specific latent factors.

\subsection{Sparsity Optimization}\label{sec:sparse}

\subsubsection{Derivation}

To alter the CorEx optimization procedure to exploit sparsity in the data, we now assume that all variables, $x_i,y_j$, are binary and $x$ is a binary vector where $X_i^\ell=1$ if word $i$ occurs in document $\ell$ and $X_i^\ell=0$ otherwise.
Since all variables are binary, the marginal distribution, $p(x_i|y_j)$, is just a two by two table of probabilities and can be estimated efficiently. 
The time-consuming part of training is the subsequent update  of the  document labels in Eq.~\ref{self consistent eqn} for each document $\ell$. 
The computation of the log likelihood ratio for all $n$ words over all documents is not efficient, as most words do not appear in a given document. We rewrite the logarithm in the interior of the sum.
\begin{align}
\lefteqn{\log \frac{p_t(x_i^\ell \mid y_j)}{p(x_i^\ell)} = \log \frac{p_t(X_i = 0 \mid y_j)}{p(X_i = 0)} +}  \label{log optimization} \\
& \qquad \quad  x_i^l  \log \left( \frac{p_t(X^\ell_i = 1 \mid y_j) p(X_i = 0)}{p_t(X_i = 0 \mid y_j) p(X_i^\ell = 1)} \right) \nonumber
\end{align}
Note, when the word does not appear in the document, only the leading term of Eq. \ref{log optimization} will be nonzero. However, when the word does appear, everything but $\log P(X_i^\ell = 1 \mid y_j) / p(X_i^\ell = 1)$ cancels out. So, we have taken advantage of the fact that the CorEx topic model binarizes documents to assume by default that a word does not appear in the document, and then correct the contribution to the update if the word does appear. 

\begin{figure}[!htb]
\includegraphics[scale = .465, trim = 0 0 0 22.5]{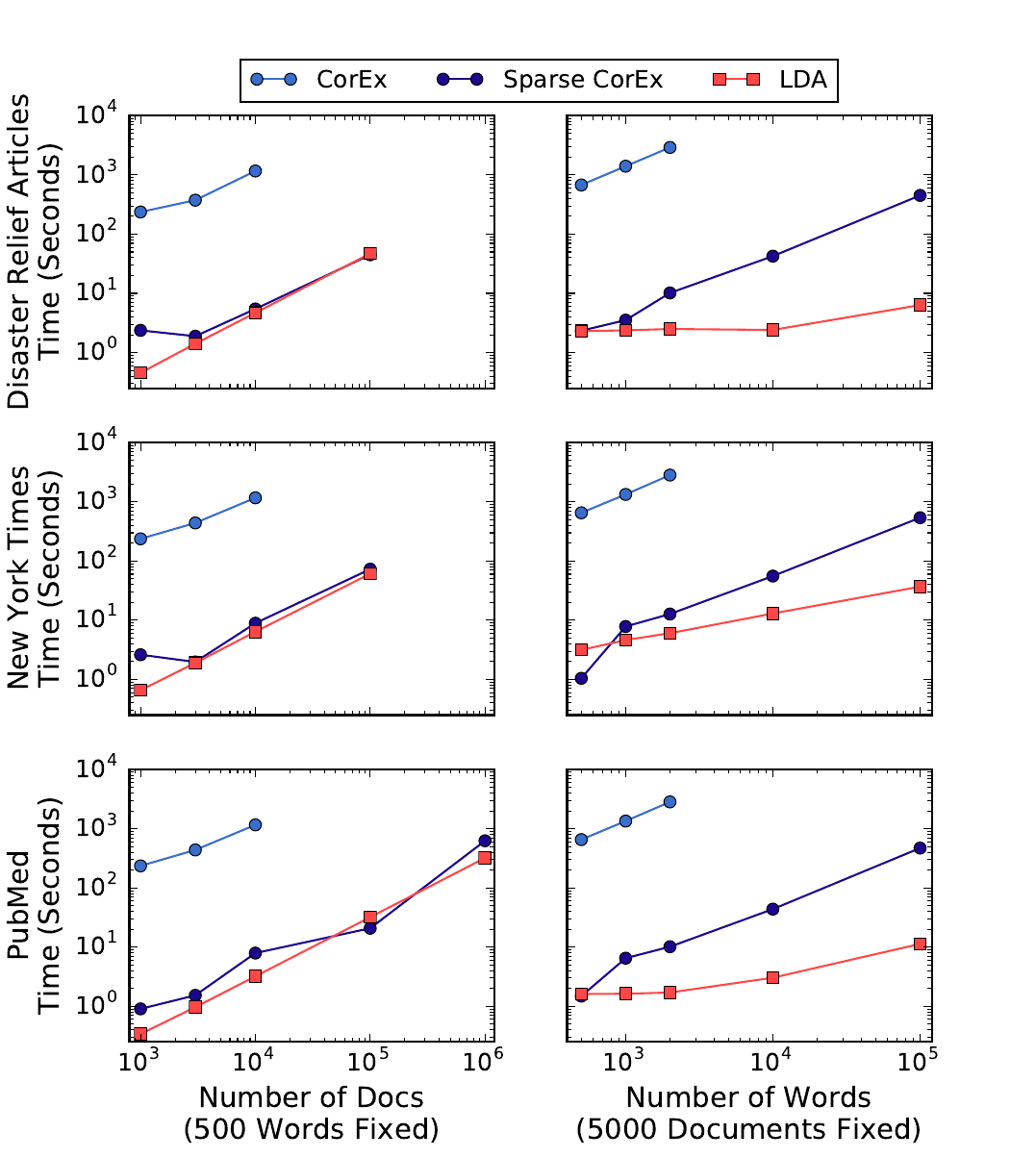}
\caption{Speed comparisons to a fixed number of iterations as the number of documents and words vary. New York Times articles and PubMed abstracts were collected from the UCI Machine Learning Repository \protect\cite{Lichman:2013}. The disaster relief articles are described in section 4.1, and represented simply as bags of words, not phrases.}
\label{speed}
\end{figure}

Thus, when substituting back into Eq.~\ref{self consistent eqn}, the sum becomes a matrix multiplication between a matrix with dimensions of the number of variables by the number of documents and entries $x_i^\ell$ that is assumed to be sparse and a dense matrix with dimensions of the number of variables by the number of latent factors. Given $n$ variables, $N$ samples, and $\rho$ nonzero entries in the data matrix, the asymptotic scaling for CorEx goes from $O(N n)$ to $O(n)+O(N) +O(\rho)$ exploiting sparsity. 
Latent tree modeling approaches are quadratic in $n$ or worse, so we expect CorEx's computational advantage to increase for larger datasets.

\subsubsection{Optimization Evaluation}

We perform experiments comparing the running time of CorEx before and after implementing the improvements which exploit sparsity.  We also compare with Scikit-Learn's simple batch implementation of LDA using the variational Bayes algorithm~\cite{hoffmanlda}.  Experiments were performed on a four core, Intel i5 chip running at 4 GHz with 32 GB RAM. We show run time when varying the data size in terms of the number of word types and the number of documents. We used 50 topics for all runs and set the number of iterations for each run to 10 iterations for LDA and 50 iterations for CorEx. 
Results are shown in Figure~\ref{speed}.  We see that CorEx exploiting sparsity is orders of magnitude faster than the naive version and is generally comparable to LDA as the number of documents scales. The slope on the log-log plot suggests a linear dependence of running time on the dataset size, as expected.


\subsection{Anchor Words via the Bottleneck}

The information bottleneck formulates a trade-off between compressing data $X$ into a representation $Y$, and preserving the information in $X$ that is relevant to $Z$ (typically labels in a supervised learning task) \cite{tishby2000information,friedman2001multivariate}. More formally, the information bottleneck is expressed as
\begin{equation}
\max_{p(y|x)} \, \beta I(Z : Y) - I(X : Y), \label{bottleneck}
\end{equation}
where $\beta$ is a parameter controlling the trade-off between compressing $X$ and preserving information about the relevance variable, $Z$. 

To see the connection with CorEx, we compare the CorEx objective as written in Eq.~\ref{almost info bottleneck} with the bottleneck in Eq.~\ref{bottleneck}. We see that we have exactly the same compression term for each latent factor, $I(X:Y_j)$, but the relevance variables now correspond to $Z \equiv X_i$. 
If we want to learn representations that are more relevant to specific keywords, we can simply anchor a word type $X_i$ to topic $Y_j$, by constraining our optimization so that $\alpha_{i,j} = \beta_{i,j}$, where $\beta_{i,j} \geq 1$ controls the anchor strength. Otherwise, the updates on $\alpha$ remain the same. This schema is a natural extension of the CorEx optimization and it is flexible, allowing for multiple word types to be anchored to one topic, for one word type to be anchored to multiple topics, or for any combination of these semi-supervised anchoring strategies.


\section{Related Work}

With respect to integrating domain knowledge into topic models, we draw inspiration from Arora et al. \shortcite{arora2012learning}, who used anchor words in the context of non-negative matrix factorization. Using an assumption of separability, these anchor words act as high precision markers of particular topics and, thus, help discern the topics from one another. Although the original algorithm proposed by Arora et al. \shortcite{arora2012learning}, and subsequent improvements to their approach, find these anchor words automatically \cite{arora2013practical,lee2014low}, recent adaptations allow manual insertion of anchor words and other metadata \cite{nguyen2014anchors,nguyen2015your}. 
Our work is similar to the latter, where we treat anchor words as fuzzy logic markers and embed them into the topic model in a semi-supervised fashion.
In this sense, our work is closest to Halpern et al. \shortcite{halpern2014using,halpern2015anchored}, who have also made use of domain expertise and semi-supervised anchored words in devising topic models.

There is an adjacent line of work that has focused on incorporating word-level information into LDA-based models. Jagarlamudi et al. \shortcite{jagarlamudi2012incorporating} proposed SeededLDA, a model that seeds words into given topics and guides, but does not force, these topics towards these integrated words. Andrzejewski and Zhu \shortcite{andrzejewski2009latent} presented a model that makes use of ``$z$-labels,'' words that are known to pertain to specific topics and that are restricted to appearing in some subset of all the possible topics. Although the $z$-labels can be leveraged to place different senses of a word into different topics, it requires additional effort to determine when these different senses occur. Our anchoring approach allows a user to more easily anchor one word to multiple topics, allowing CorEx to naturally find topics that revolve around different senses of a word.

Andrzejewski et al.~\shortcite{andrzejewski2009incorporating} presented a second model which allows specification of Must-Link and Cannot-Link relationships between  words that help partition otherwise muddled topics. These logical constraints help enforce topic separability, though these mechanisms less directly address how to anchor a single word or set of words to help a topic emerge. More generally, the Must/Cannot link and $z$-label topic models have been expressed in a powerful first-order-logic framework that allows the specification of arbitrary domain knowledge through logical rules \cite{andrzejewski2011framework}. Others have built off this first-order-logic approach to automatically learn rule weights \cite{mei2014robust} and incorporate additional latent variable information \cite{foulds2015latent}.

Mathematically, CorEx topic models most closely resemble topic models based on latent tree reconstruction~\cite{chen_topic}. In Chen et al.'s \shortcite{chen_topic} analysis, their own latent tree approach and CorEx both report significantly better perplexity than hierarchical topic models based on the hierarchical Dirichlet process and the Chinese restaurant process. CorEx has also been investigated as a way to find ``surprising'' documents~\cite{hodas}.
\nocite{icml_interpret}


\section{Data and Evaluation Methods}

\subsection{Data}

We use two challenging datasets with corresponding domain knowledge lexicons to evaluate anchored CorEx. Our first dataset consists of 504,000 humanitarian assistance and disaster relief (HA/DR) articles covering 21 disaster types collected from ReliefWeb, an HA/DR news article aggregator sponsored by the United Nations \cite{Littell2018}. To mitigate overwhelming label imbalances during anchoring, we both restrict ourselves to documents in English with one label, and randomly subsample 2,000 articles from each of the largest disaster type labels. This leaves us with a corpus of 18,943 articles.\footnote{HA/DR articles and accompanying lexicon available at \url{http://dx.doi.org/10.7910/DVN/TGOPRU}}

We accompany these articles with an HA/DR lexicon of approximately 34,000 words and phrases \cite{Littell2018}. The lexicon was curated by first gathering 40--60 seed terms per disaster type from HA/DR domain experts and CrisisLex. This term list was then expanded by creating word embeddings for each disaster type, and taking terms within a specified cosine similarity of the seed words. These lists were then filtered by removing names, places, non-ASCII characters, and terms with fewer than three characters. Finally, the extracted terms were audited using CrowdFlower, where users rated the relevance of the terms on a Likert scale. Low relevance terms were dropped from the lexicon. Of these terms 11,891 types appear in the HA/DR articles.

Our second dataset consists of 1,237 deidentified clinical discharge summaries from the Informatics for Integrating Biology and the Bedside (i2b2) 2008 Obesity Challenge.\footnote{Data available upon data use agreement at \url{https://www.i2b2.org/NLP/Obesity/}} These summaries are labeled by clinical experts with 15 conditions frequently associated with obesity. For these documents, we leverage a text pipeline that extracts common medical terms and phrases \cite{dai2008efficient,chapman2001simple}, which yields 3,231 such term types. For both sets of documents, we use their respective lexicons to break the documents down into bags of words and phrases.

We also make use of the 20 Newsgroups dataset, as provided and preprocessed in the Scikit-Learn library \cite{scikit-learn}.


\subsection{Evaluation}

CorEx does not explicitly attempt to learn a generative model and, thus, traditional measures such as perplexity are not appropriate for model comparison against LDA. Furthermore, it is well-known that perplexity and held-out log-likelihood do not necessarily correlate with human evaluation of semantic topic quality \cite{chang2009reading}. Therefore, we measure the semantic topic quality using Mimno et al.'s \shortcite{mimno2011optimizing} UMass automatic topic coherence score, which correlates with human judgments.

We also evaluate the models in terms of multiclass logistic regression document classification \cite{scikit-learn}, where the feature set of each document is its topic distribution. 
We perform all document classification tasks using a 60/40 training-test split.

Finally, we measure how well each topic model does at clustering documents. We obtain a clustering by assigning each document to the topic that occurs with the highest probability. 
We then measure the quality within clusters (homogeneity) and across clusters (adjusted mutual information). The highest possible value for both measures is one. We do not report clustering metrics on the clinical health notes because the documents are multi-label and, in that case, the metrics are not well-defined.

\subsection{Choosing Anchor Words}


We follow the approach used by Jagarlamudi et al. \shortcite{jagarlamudi2012incorporating} to automatically generate anchor words: for each label in a data set, we find the words that have the highest mutual information with the label. For word $w$ and label $L$, this is computed as
\begin{equation}
I(L : w) = H(L) - H(L \mid w),
\end{equation}
where for each document of label $L$ we consider if the word $w$ appears or not.

\begin{figure}[!th]
\includegraphics[scale = .47, trim = 0 0 0 15]{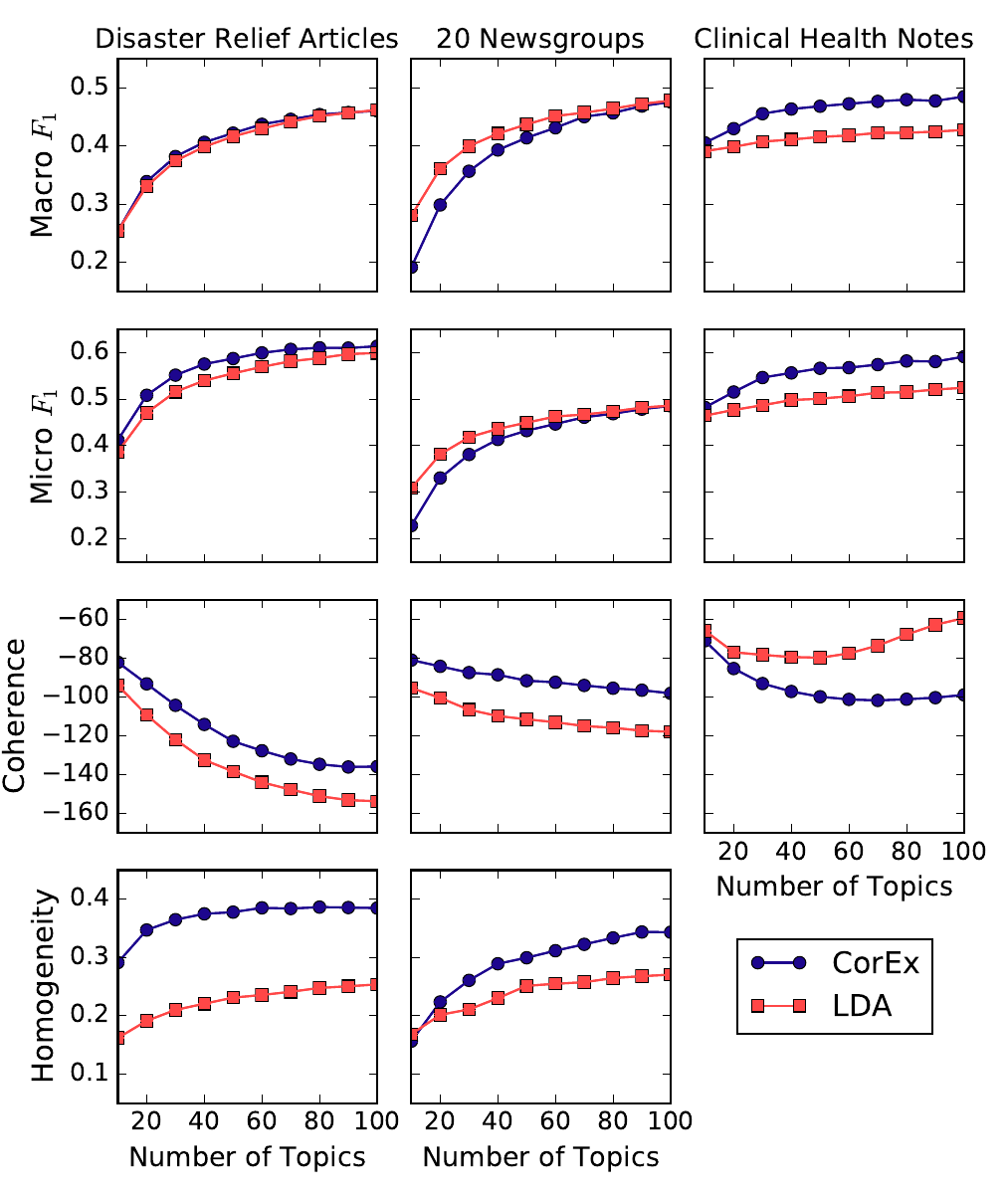}
\caption{Baseline comparison of CorEx to LDA with respect to topic coherence and document classification and clustering on three different datasets as the number of topics vary. Points are the average of 30 runs of a topic model. Confidence intervals are plotted but are so small that they are not distinguishable. CorEx is trained using binary data, while LDA is trained on count data. Homogeneity is not well-defined on the multi-label clinical health notes, so it is omitted.}
\label{baseline-comparison}
\end{figure}

\section{Results}

\subsection{LDA Baseline Comparison}

We compare CorEx to LDA in terms of topic coherence, document classification, and document clustering across three datasets. CorEx is trained on binary data, while LDA is trained on count data. While not reported here, CorEx consistently outperformed LDA trained on binary data. In doing these comparisons, we use the Gensim implementation of LDA \cite{rehurek_lrec}. The results of comparing CorEx to LDA as a function of the number of topics are presented in Figure~\ref{baseline-comparison}.

Across all three datasets, we find that the topics produced by CorEx yield document classification results that are on par with or better than those produced by LDA topics. In terms of clustering, CorEx consistently produces document clusters of higher homogeneity than LDA. On the disaster relief articles, the CorEx clusters are nearly twice as homogeneous as the LDA clusters.

\begin{table}[!tb]
\centering
\begin{tabular}{|c|l|}
\hline
\textbf{Rank} & \multicolumn{1}{c|}{\textbf{Disaster Relief Topic}} \\ \hline
1             & \specialcell{drought, farmers, harvest, crop,\\ livestock, planting, grain, maize,\\ rainfall, irrigation}                   \\ \hline
3             & \specialcell{eruption, volcanic, lava, crater,\\ eruptions, volcanos, slopes, volcanic\\ activity, evacuated, lava flows}                  \\ \hline
8             & \specialcell{winter, snow, snowfall, temperatures, \\heavy snow, heating, freezing, warm\\ clothing, severe winter, avalanches}                         \\ \hline
23            & \specialcell{military, armed, civilians, soldiers, \\aircraft, weapons, rebel, planes, bombs,\\ military personnel}                         \\ \hline
\textbf{Rank} & \multicolumn{1}{c|}{\textbf{20 Newsgroups Topic}}            \\ \hline
3             & \specialcell{team, game, season, player, league,\\ hockey, play, teams, nhl}                           \\ \hline
14            & \specialcell{car, bike, cars, engine, miles, road,\\ ride, riding, bikes, ground}                        \\ \hline
26            & \specialcell{nasa, launch, orbit, shuttle, mission,\\ satellite, gov, jpl, orbital, solar} \\ \hline
39            & \specialcell{medical, disease, doctor, patients,\\ treatment, medicine, health, hospital,\\ doctors, pain}                              \\ \hline 
\textbf{Rank} & \multicolumn{1}{c|}{\textbf{Clinical Health Notes Topic}}    \\ \hline
12             & \specialcell{vomiting, nausea, abdominal pain,\\ diarrhea, fever, dehydration, chill, \\clostridium difficile, intravenous fluid,\\ compazine}                   \\ \hline
19            & \specialcell{anxiety state, insomnia, ativan,\\ neurontin, depression, lorazepam,\\ gabapentin, trazodone, fluoxetine,\\ headache}                  \\ \hline
27            & \specialcell{pain, oxycodone, tylenol, percocet, \\ibuprofen, morphine, osteoarthritis,\\ hernia, motrin, bleeding}                           \\ \hline
\end{tabular}
\caption{Examples of topics learned by the CorEx topic model. Words are ranked according to mutual information with the topic, and topics are ranked according to the amount of total correlation they explain. 
Topic models were run with 50 topics on the Reliefweb and 20 Newsgroups datasets, and 30 topics on the clinical health notes.
}
\label{unsupervised-topics}
\end{table}

CorEx outperforms LDA in terms of topic coherence on two out of three of the datasets. While LDA produces more coherent topics for the clinical health notes, it is particularly striking that CorEx is able to produce high quality topics while only leveraging binary count data. Examples of these topics are shown in Table~\ref{unsupervised-topics}. Despite the binary counts limitation, CorEx still finds meaningfully coherent and competitive structure in the data.

\begin{figure}[!tb]
\includegraphics[scale = .465, trim = 15 10 0 0]{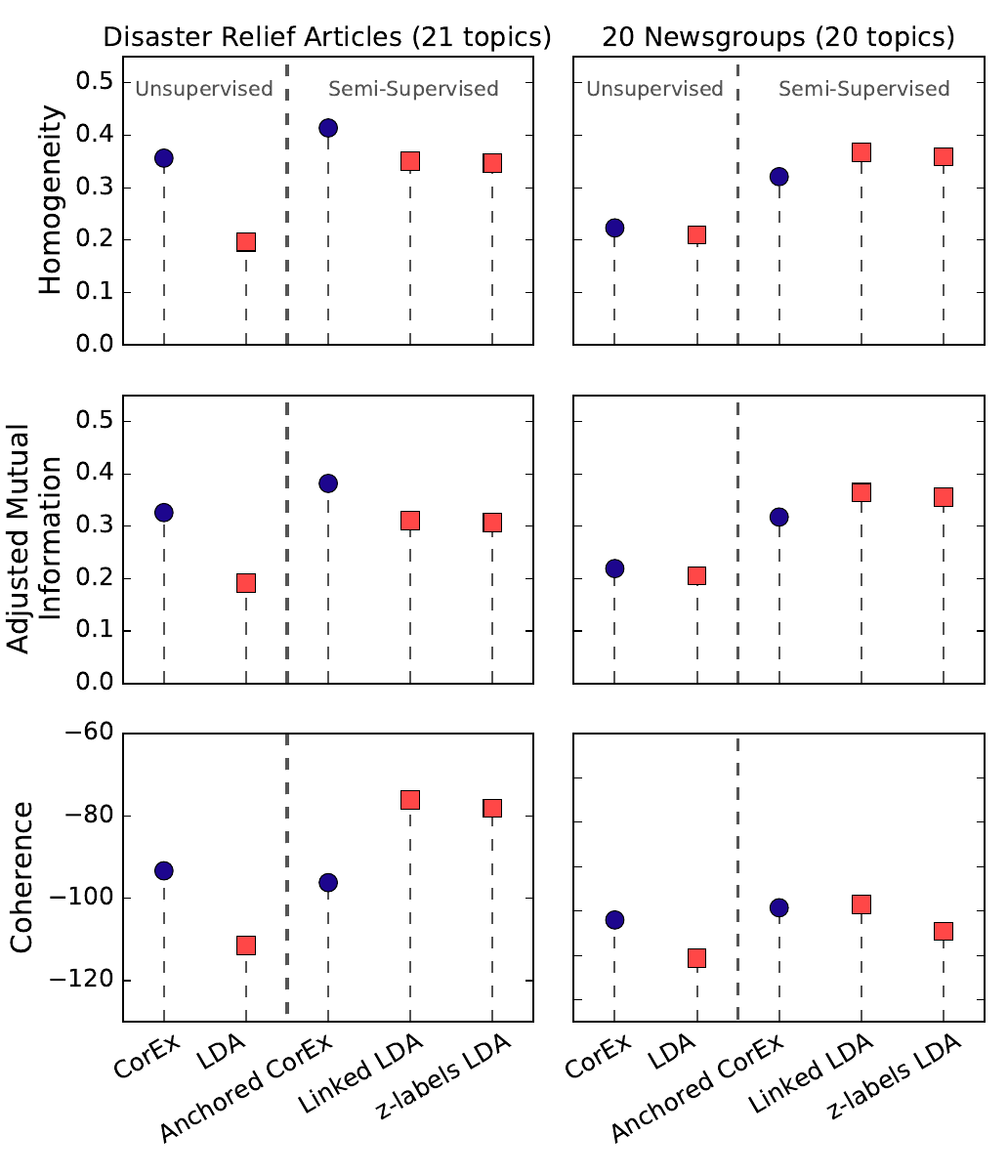}
\caption{Comparison of anchored CorEx to other semi-supervised topic models in terms of document clustering and topic coherence. For each dataset, the number of topics is fixed to the number of document labels. Each dot is the average of 30 runs. Confidence intervals are plotted but are so small that they are not distinguishable.}
\label{semi-supervised}
\end{figure}

\subsection{Anchored CorEx Analysis}

We now examine the effects and benefits of guiding CorEx through anchor words. In doing so, we also compare anchored CorEx to other semi-supervised topic models. 

\subsubsection{Anchoring for Topic Separability}

We are first interested in how anchoring can be used to encourage topic separability so that documents cluster well. We focus on the HA/DR articles and 20 newsgroups datasets, since traditional clustering metrics are not well-defined on the multi-label clinical health notes. For both datasets, we fix the number of topics to be equal to the number of document labels. It is in this context that we compare anchored CorEx to two other semi-supervised topic models: $z$-labels LDA and must/cannot link LDA.

\begin{table}[!tb]
\centering
\begin{tabular}{|c|l|}
\hline
\textbf{Rank} & \multicolumn{1}{c|}{\textbf{Anchored Disaster Relief Topic}}      \\ \hline
1             & \specialcell{\textbf{harvest}, \textbf{locus}, drought, \textbf{food crisis},\\ farmers, crops, crop, malnutrition,\\ food aid, livestock}                  \\ \hline
4             & \specialcell{\textbf{tents}, \textbf{quake}, international federation,\\ red crescent, red cross, blankets, \\earthquake, \textbf{richter scale}, societies, \\aftershocks} \\ \hline
12             & \specialcell{\textbf{climate}, \textbf{impacts}, \textbf{warming}, climate \\ change, irrigation, consumption, \\household, droughts, livelihoods,\\ interventions}                   \\ \hline
19            & \specialcell{\textbf{storms}, weather, winds, coastal,\\ \textbf{tornado}, meteorological, \textbf{tornadoes}, \\strong winds, tropical, roofs}                       \\ \hline 
\textbf{Rank} & \multicolumn{1}{c|}{\textbf{Anchored 20 Newsgroups Topic}}                 \\ \hline
5             & \specialcell{government, \textbf{congress}, \textbf{clinton}, state,\\ national, economic, general, states,\\ united, order}                  \\ \hline
6             & \specialcell{\textbf{bible}, \textbf{christian}, god, jesus, christians,\\ believe, life, faith, world, man}                     \\ \hline
15            & \specialcell{use, used, high, \textbf{circuit}, power, work, \\voltage, need, low, end }                                 \\ \hline
20            & \specialcell{\textbf{baseball}, \textbf{pitching}, \textbf{braves}, \textbf{mets}, \\ \textbf{hitter}, pitcher, cubs, dl, sox, jays}                     \\ \hline
\end{tabular}
\caption{Examples of topics learned by CorEx when simultaneously anchoring many topics with anchoring parameter $\beta = 2$. Anchor words are shown in \textbf{bold}. Words are ranked according to mutual information with the topic, and topics are ranked according to the amount of total correlation they explain. Topic models were run with 21 topics on the Reliefweb articles and 20 topics on the 20 Newsgroups dataset.}
\label{anchored-topics}
\end{table}

Using the method described in Section 4.3, we automatically retrieve the top five anchors for each disaster type and newsgroup. We then filter these lists of any words that are ambiguous, i.e. words that are anchor words for more than one document label. For anchored CorEx and $z$-labels LDA we simultaneously assign each set of anchor words to exactly one topic each. For must/cannot link LDA, we create must-links within the words of the same anchor group, and create cannot-links between words of different anchor groups.

Since we are simultaneously anchoring to many topics, we use a weak anchoring parameter $\beta = 2$ for anchored CorEx. Using the notation from their original papers, we use $\eta = 1$ for $z$-labels LDA, and $\eta = 1000$ for must/cannot link LDA. For both LDA variants, we use $\alpha = 0.5$, $\beta = 0.1$ and take 2,000 samples, and estimate the models using code implemented by the original authors.

The results of this comparison are shown in Figure~\ref{semi-supervised}, and examples of anchored CorEx topics are shown in Table~\ref{anchored-topics}. Across all measures CorEx and anchored CorEx outperform LDA. We find that anchored CorEx always improves cluster quality versus CorEx in terms of homogeneity and adjusted mutual information. Compared to CorEx, multiple simultaneous anchoring neither harms nor benefits the topic coherence of anchored CorEx. Together these metrics suggest that anchored CorEx is finding topics that are of equivalent coherence to CorEx, but more relevant to the document labels since gains are seen in terms of document clustering.

Against the other semi-supervised topic models, anchored CorEx compares favorably. The document clustering of anchored CorEx is similar to, or better than, that of $z$-labels LDA and must/cannot link LDA. Across the disaster relief articles, anchored CorEx finds less coherent topics than the two LDA variants, while it finds similarly coherent topics as must/cannot link LDA on the 20 newsgroup dataset.

\subsubsection{Anchoring for Topic Representation}

We now turn to studying how domain knowledge can be anchored to a single topic to help an otherwise dominated topic emerge, and how the anchoring parameter $\beta$ affects that emergence. To discern this effect, we focus just on anchored CorEx along with the HA/DR articles and clinical health notes, datasets for which we have a domain expert lexicon.

We devise the following experiment: first, we determine the top five anchor words for each document label using the methodology described in Section 4.3. Unlike in the previous section, we do not filter these lists of ambiguous anchor words. Second, for each document label, we run an anchored CorEx topic model with that label's anchor words anchored to exactly one topic. We compare this anchored topic model to an unsupervised CorEx topic model using the same random seeds, thus creating a matched pair where the only difference is the treatment of anchor words. Finally, this matched pairs process is repeated 30 times, yielding a distribution for each metric over each label.

\begin{figure}[!t]
\includegraphics[scale = .45, trim = 0 0 0 0]{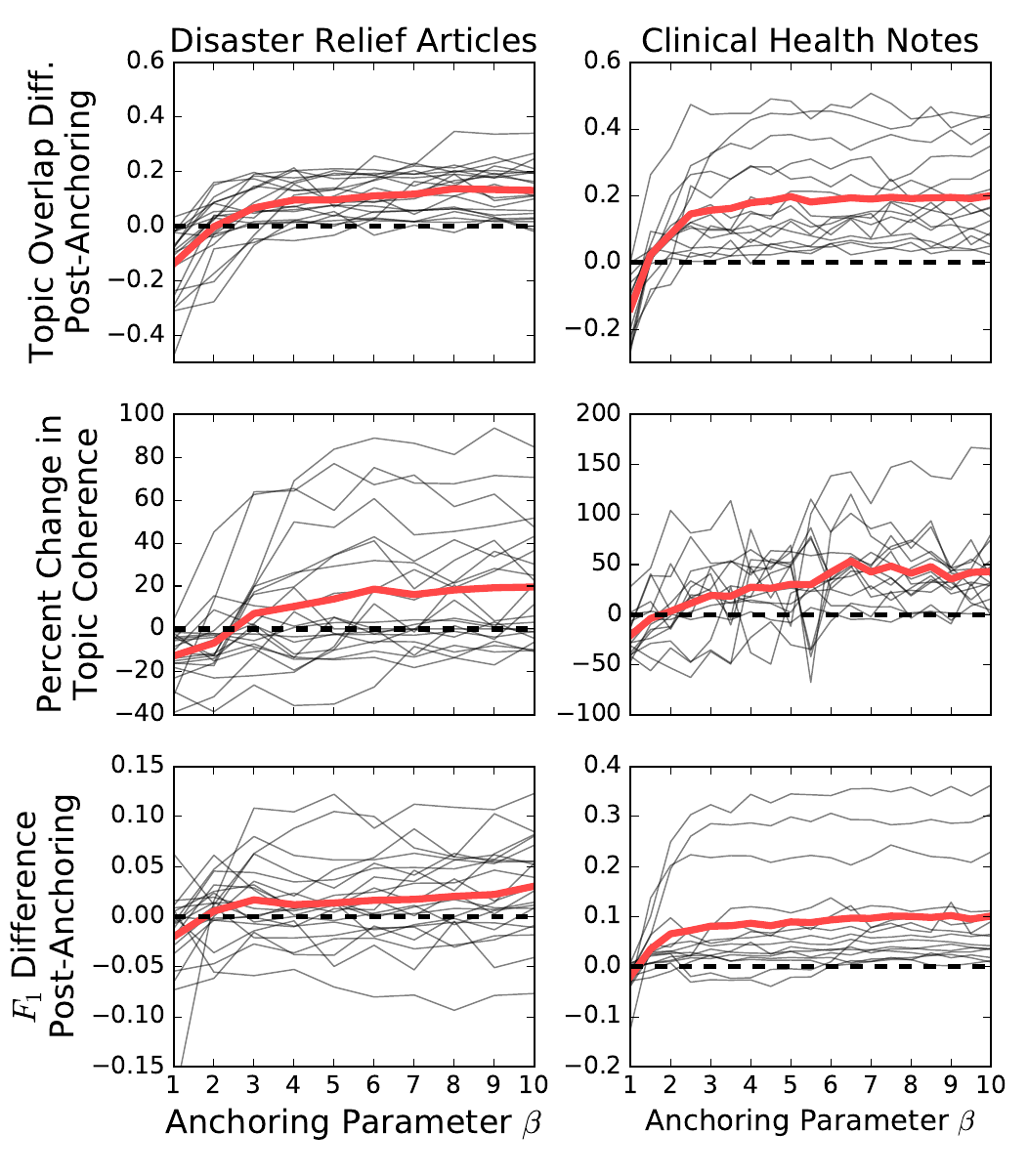}
\caption{Effect of anchoring words to a single topic for one document label at a time as a function of the anchoring parameter $\beta$. Light gray lines indicate the trajectory of the metric for a given disaster or disease label. Thick red lines indicate the pointwise average across all labels for fixed value of $\beta$.}
\label{varying-beta}
\end{figure}

\begin{figure*}[!htbp]
\centering
\includegraphics[scale = .38, trim = 30 0 0 0]{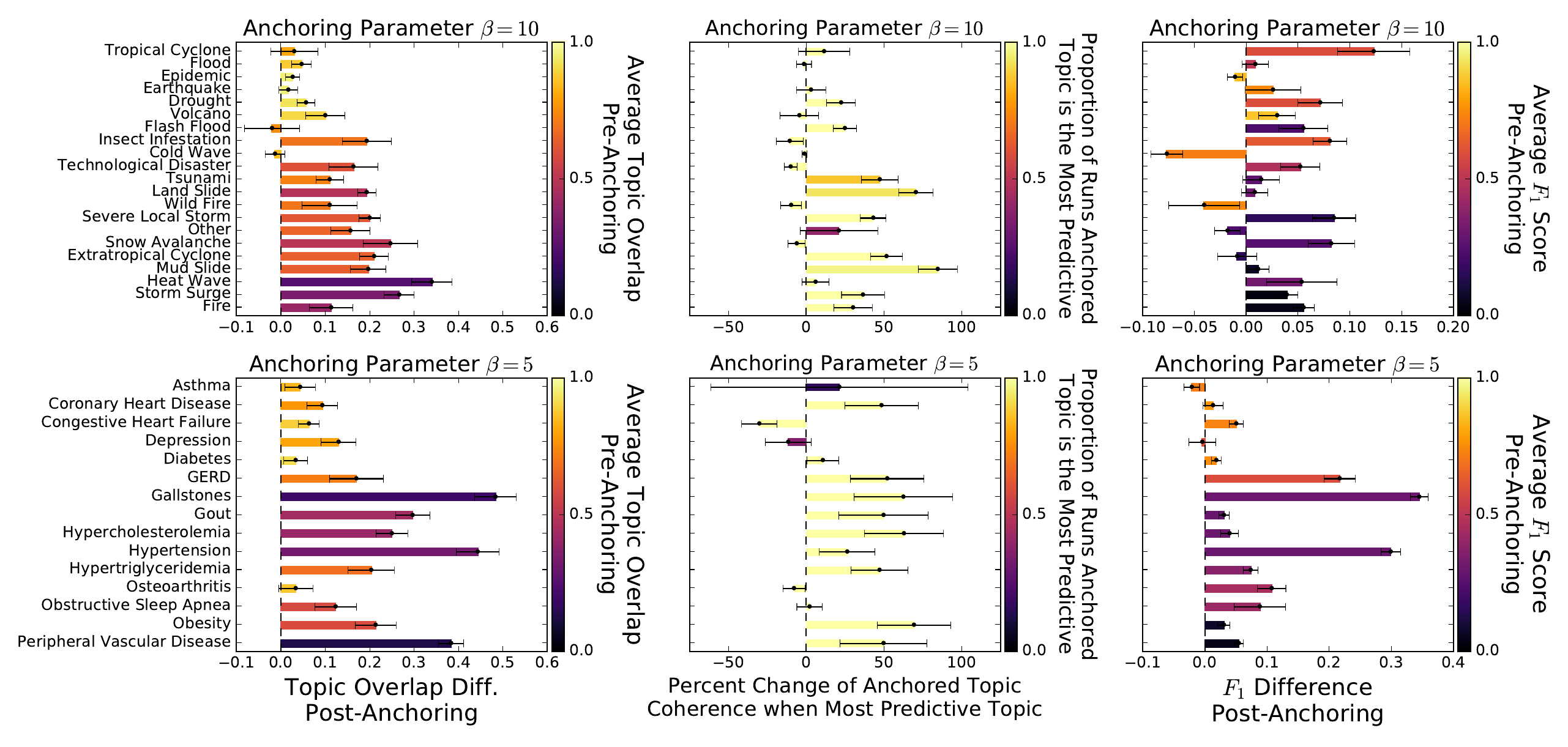}
\caption{Cross-section results of the anchoring metrics from fixing $\beta = 5$ for the clinical health notes, and $\beta = 10$ for the disaster relief articles. Disaster and disease types are sorted by frequency, with the most frequent document labels appearing at the top. Error bars indicate 95\% confidence intervals. The color bars provide context for each metric: topic overlap pre-anchoring, proportion of topic model runs where the anchored topic was the most predictive topic, and $F_1$ score pre-anchoring.}
\label{fixed-beta}
\end{figure*}

We use 50 topics when modeling the ReliefWeb articles and 30 topics when modeling the i2b2 clinical health notes. These values were chosen by observing diminishing returns to the total correlation explained by additional topics.

In Figure~\ref{varying-beta} we show how the results of this experiment vary as a function of the anchoring parameter $\beta$ for each disaster and disease type in the two data sets. Since there is heavy variance across document labels for each metric, we also examine a more detailed cross section of these results in Figure~\ref{fixed-beta}, where we set $\beta = 5$ for the clinical health notes and set $\beta = 10$ for the disaster relief articles. As we show momentarily, disaster and disease types that benefit the most from anchoring were underrepresented pre-anchoring. Document labels that were well-represented prior to anchoring achieve only marginal gain. This results in the variance seen in Figure~\ref{varying-beta}.

A priori we do not know that anchoring will cause the anchor words to appear at the top of topics. So, we first measure how the topic overlap, the proportion of the top ten mutual information words that appear within the top ten words of the topics, changes before and after anchoring. From Figure~\ref{varying-beta} (row 1) we see that as $\beta$ increases, more of these relevant words consistently appear within the topics. For the disaster relief articles, many disaster types see about two more words introduced, while in the clinical health notes the overlap increases by up to four words. Analyzing the cross section in Figure~\ref{fixed-beta} (column 1), we see many of these gains come from disaster and disease types that appeared less in the topics pre-anchoring. Thus, we can sway the topic model towards less dominant themes through anchoring. Document labels that occur the most frequently are those for which the topic overlap changes the least.

Next, we examine whether these anchored topics are more coherent topics. To do so, we compare the coherence of the anchored topic with that of the most predictive topic pre-anchoring, i.e. the topic with the largest corresponding coefficient in magnitude of the logistic regression, when the anchored topic itself is most predictive. From Figure~\ref{varying-beta} (row 2), we see these results have more variance, but largely the anchored topics are more coherent. In some cases, the coherence is 1.5 to 2 times that of pre-anchoring. Furthermore, by colors of the central panel of Figure~\ref{fixed-beta}, we find that the anchored topics are, indeed, often the most predictive topics for each document label. Similar to topic overlap, the labels that see the least improvement are those that appear the most and are already well-represented in the topic model.

Finally, we find that the anchored, more coherent topics can lead to modest gains in document classification. For the disaster relief articles, Figure~\ref{varying-beta} (row 3) shows that there are mixed results in terms of $F_1$ score improvement, with some disaster types performing consistently better, and others performing consistently worse. The results are more consistent for the clinical health notes, where there is an average increase of about 0.1 in the $F_1$ score, and some disease types see an increase of up to 0.3 in $F_1$. Given that we are only anchoring 5 words to the topic model, these are significant gains in predictive power.

Unlike the gains in topic overlap and coherence, the $F_1$ score increases do not simply correlate with which document labels appeared most frequently. For example, we see in Figure~\ref{fixed-beta} (column 3) that Tropical Cyclone exhibits the largest increase in predictive performance, even though it is also one of the most frequently appearing document labels. Similarly, some of the major gains in $F_1$ for the disease types, and major losses in $F_1$ for the disaster types, do not come from the most or least frequent document labels. Thus, if using anchoring single topics within CorEx for document classification, it is important to examine how the anchoring affects prediction for individual document labels.


\subsubsection{Anchoring for Topic Aspects}

Finding topics that revolve around a word, such as a name or location, or a group of words can aid in understanding how a particular subject or event has been framed. We finish with a qualitative experiment where we disambiguate aspects of a topic by anchoring a set of words to multiple topics within the CorEx topic model. Note, must/cannot link LDA cannot be used in this manner, and $z$-labels LDA would require us to know these aspects beforehand.

We consider tweets containing \#Ferguson (case-insensitive), which detail reactions to the shooting of Black teenager Michael Brown by White police officer Darren Wilson on August 9th, 2014 in Ferguson, Missouri. These tweets were collected from the Twitter Gardenhose, a 10\% random sample of all tweets, over the period August 9th, 2014 to November 30th, 2014. Since CorEx will seek maximally informative topics by exploiting redundancies, we remove duplicates of retweets, leaving us with 869,091 tweets. We filter these tweets of punctuation, stop words, hyperlinks, usernames, and the `RT' retweet symbol, and use the top 20,000 word types.

In the wake of both the shooting and the eventual non-indictment of Darren Wilson, several protests occurred. Some onlookers supported and encouraged such protests, while others characterized the protests as violent ``riots." To disambiguate these different depictions, we train a CorEx topic model with 55 topics, anchoring ``protest" and ``protests" together to five topics, and ``riot" and ``riots" together to five topics with $\beta = 2$. These anchored topics are presented in Table~\ref{aspect-topics}.

\begin{table}[t]
\centering
\begin{tabular}{|c|l|}
\hline
\multicolumn{2}{|c|}{\textbf{Topic Aspects of ``protest"}}                                             \\ \hline
1  & \specialcell{\textbf{protest}, \textbf{protests}, peaceful, violent, continue,\\ night, island, photos, staten, nights}       \\ \hline
2  & \specialcell{\textbf{protest}, \textbf{protests}, \#hiphopmoves, \#cole,\\ hiphop, nationwide, moves, fo, anheuser,\\ boeing}   \\ \hline
3  & \specialcell{\textbf{protest}, \textbf{protests}, st, louis, guard, national,\\ county, patrol, highway, city }               \\ \hline
4  & \specialcell{\textbf{protest}, \textbf{protests}, paddy, covering, beverly,\\ walmart, wagon, hills, passionately,\\ including} \\ \hline
5  & \specialcell{\textbf{protest}, \textbf{protests}, solidarity, march, square,\\ rally, \#oakland, downtown, nyc, \#nyc}        \\ \hline \hline
\multicolumn{2}{|c|}{\textbf{Topic Aspects of ``riot"}}                                                    \\ \hline
6  & \specialcell{\textbf{riot}, \textbf{riots}, unheard, language, inciting,\\ accidentally, jokingly, watts, waving, dies}       \\ \hline
7  & \specialcell{\textbf{riot}, black, \textbf{riots}, white, \#tcot, blacks, men,\\ whites, race, \#pjnet}                       \\ \hline
8  & \specialcell{\textbf{riot}, \textbf{riots}, looks, like, sounds, acting, act,\\ animals, looked, treated}                     \\ \hline
9  & \specialcell{\textbf{riot}, \textbf{riots}, store, looting, businesses,\\ burning, fire, looted, stores, business}            \\ \hline
10 & \specialcell{gas, \textbf{riot}, tear, \textbf{riots}, gear, rubber, bullets,\\ military, molotov, armored}                   \\ \hline
\end{tabular}
\caption{Topic aspects around ``protest" and ``riot" from running a CorEx topic model with 55 topics and anchoring ``protest" and ``protests" together to five topics and ``riot" and ``riots" together to five topics with $\beta = 2$. Anchor words are shown in \textbf{bold}. Note, topics are not ordered by total correlation.}
\label{aspect-topics}
\end{table}

The anchored topics reflect different aspects of the framing of the ``protests" and ``riots," and are generally interpretable, despite the typical difficulty of extracting coherent topics from short documents using LDA \cite{tang2014understanding}. 
The ``protest" topic aspects describe protests in St. Louis, Oakland, Beverly Hills, and parts of New York City (topics 1, 3, 4, 5), resistance by law enforcement (topics 3 and 4), and discussion of whether the protests were peaceful (topic 1). Topic 2 revolves around hip-hop artists who marched in solidarity with protesters. The ``riot" topic aspects discuss racial dynamics of the protests (topic 7) and suggest the demonstrations are dangerous (topics 8 and 9). Topic 10 describes the ``riot" gear used in the militarized response to the Ferguson protesters, and Topic 7 also hints at aspects of conservatism through the hashtags \#tcot (Top Conservatives on Twitter) and \#pjnet (Patriot Journalist Network).

As we see, anchored CorEx finds several interesting, non-trivial aspects around ``protest" and ``riot" that could spark additional qualitative investigation. Retrieving topic aspects through anchor words in this manner allows the user to explore different frames of complex issues, events, or discussions within documents. As with the other anchoring strategies, this has the potential to supplement qualitative research done by researchers within the social sciences and digital humanities.


\section{Discussion}

We have introduced an information-theoretic topic model, CorEx, that does not rely on any of the generative assumptions of LDA-based topic models. This topic model seeks maximally informative topics as encoded by their total correlation. We also derived a flexible method for anchoring word-level domain knowledge in the CorEx topic model through the information bottleneck. Anchored CorEx guides the topic model towards themes that do not naturally emerge, and often produces more coherent and predictive topics. Both CorEx and anchored CorEx consistently produce topics that are of comparable quality to LDA-based methods, despite only making use of binarized word counts.

Anchored CorEx is more flexible than previous attempts at integrating word-level information into topic models. Topic separability can be enforced by lightly anchoring disjoint groups of words to separate topics, topic representation can be promoted by assertively anchoring a group of words to a single topic, and topic aspects can be unveiled by anchoring a single group of words to multiple topics. The flexibility of anchoring through the information bottleneck lends itself to many other possible creative anchoring strategies that could guide the topic model in different ways. Different goals may call for different anchoring strategies, and domain experts can shape these strategies to their needs.

While we have demonstrated several advantages of the CorEx topic model to LDA, it does have some technical shortcomings. Most notably, CorEx relies on binary count data in its sparsity optimization, rather than the standard count data that is used as input into LDA and other topic models. While we have demonstrated CorEx performs at the level of LDA despite this limitation, its effect would be more noticeable on longer documents. This can be partly overcome if one chunks such longer documents into shorter subdocuments prior to running the topic model. Our implementation also requires that each word appears in only one topic. These limitations are not fundamental limitations of the theory, but a matter of computational efficiency. In future work, we hope to remove these restrictions while preserving the speed of the sparse CorEx topic modeling algorithm.

As we have demonstrated, the information-theoretic approach provided via CorEx has rich potential for finding meaningful structure in documents, particularly in a way that can help domain experts guide topic models with minimal intervention to capture otherwise eclipsed themes. The lightweight and versatile framework of anchored CorEx leaves open possibilities for theoretical extensions and novel applications within the realm of topic modeling.

\section*{Acknowledgments}

We would like to thank the Machine Intelligence and Data Science (MINDS) research group at the Information Sciences Institute for their help and insight during the course of this research. We also thank the Vermont Advanced Computing Core (VACC) for its computational resources. We acknowledge the construction of the HA/DR corpus and lexicon by Leidos Corp. under funding from the Defense Advanced Research Projects Agency (DARPA) Information Innovation Office (I2O), program: Low Resource Languages for Emergent Incidents (LORELEI), Contract No. HR0011-15-C-0114. Finally, we thank the anonymous reviewers and the TACL action editors Diane McCarthy and Kristina Toutanova for their time and effort in helping us improve our work. Ryan J. Gallagher was a visiting research assistant at the Information Sciences Institute while performing this research. Ryan J. Gallagher and Greg Ver Steeg were supported by DARPA award HR0011-15-C-0115 and David Kale was supported by the Alfred E. Mann Innovation in Engineering Doctoral Fellowship.

\bibliography{bibl}
\bibliographystyle{acl2012}

\clearpage

\appendix

\renewcommand\thefigure{\thesection\arabic{figure}}  
\renewcommand\thetable{\thesection\arabic{table}}

\setcounter{figure}{0} 
\setcounter{table}{0}

\section{Supplemental Material: Anchor Words and Topic Examples}
\label{sec:supplemental}

\begin{table}[!th]
\small
\begin{center}
\begin{tabular}{|l|l|}
\hline
\textbf{{\normalsize Disease Type}}         & \textbf{{\normalsize Anchor Words}}                                         \\ \hline
{\normalsize Asthma}                                     & \specialcell{asthma, albuterol, wheeze,\\ advair, fluticasone}                                                           \\ \hline
\specialcell{{\normalsize Coronary Artery}\\{\normalsize Disease}}                    & \specialcell{coronary artery disease,\\ aspirin, myocardial\\ inarction, plavix}                                           \\ \hline
\specialcell{{\normalsize Congestive Heart}\\ {\normalsize Failure}}                   & \specialcell{congestive heart failure,\\ lasix, diuresis, heart failure,\\cardiomyopathy}                                \\ \hline
{\normalsize Depression}                                 & \specialcell{depression, prozac, celexa,\\ seroquel, remeron}                                                            \\ \hline
{\normalsize Diabetes}                                   & \specialcell{diabetes mellitus, diabetes,\\ nph insulin, insulin,\\ metformin}                                             \\ \hline
\specialcell{{\normalsize Gastroesophageal}\\{\normalsize Reflux Disease}}          & \specialcell{gastroesophageal refulx,\\ no known drug allergy,\\protonix, not:, reflux}                                   \\ \hline
{\normalsize Gallstones}                                 & \specialcell{gallstone, cholecystitis,\\ cholelithiasis,\\ abdominal pain,vomiting}                                       \\ \hline
{\normalsize Gout}                                       & \specialcell{gout, allopurinol,\\ colchicine, renal\\ insufficiency, torsemide}                                            \\ \hline
{\normalsize Hypercholesterolemia}                      & \specialcell{hypercholesterolemia,\\hyperlipidemia, aspirin,\\ lipitor, dyslipidemia}                                     \\ \hline
{\normalsize Hypertension}                               & \specialcell{hypertension, lisinopril,\\ aspirin, diabetes mellitus,\\ atorvastatin}                                       \\ \hline
{\normalsize Hypertriglyceridemia}                       & \specialcell{{\small hypertriglyceridemia,}\\gemfibrozil,\\ citrate, orphenadrine,\\ hydroxymethylglutaryl coa\\reductase inhibitors} \\ \hline
{\normalsize Osteoarthritis}                             & \specialcell{osteoarthritis, degenerative\\ joint disease, arthritis,\\ naproxen, fibromyalgia}                            \\ \hline
\specialcell{{\normalsize Obstructive}\\{\normalsize Sleep Apnea}}                    & \specialcell{sleep apnea, obstructive\\ sleep apnea, morbid obese,\\ obesity, ipratropium}                                 \\ \hline
{\normalsize Obesity}                                    & \specialcell{obesity, morbid obesity,\\ obese, sleep apnea,\\ coronary artery disease}                                     \\ \hline
\specialcell{{\normalsize Peripheral Vascular}\\ {\normalsize Disease}}                & \specialcell{cellulitis, erythema,\\ulcer, swelling,\\word finding difficulty}  \\ \hline     
\end{tabular}
\end{center}
\caption{Words that have the highest mutual information with each disease type.}
\label{disease-anchor-words}
\end{table}

\begin{table}[!th]
\normalsize
\begin{center}
\begin{tabular}{|l|l|}
\hline
\textbf{Disaster Type}         & \textbf{Anchor Words}                                         \\ \hline
Cold Wave                      & \specialcell{winter, snow, cold,\\ temperatures, heavy snow}                  \\ \hline
Drought                        & \specialcell{drought, taliban, wheat,\\ refugees, severe drought}             \\ \hline
Earthquake                     & \specialcell{earthquake, quake,\\ richter scale, tents, injured}              \\ \hline
Epidemic                       & \specialcell{virus, ebola outbreak,\\ transmission, ebola virus,\\ disaster}    \\ \hline
\specialcell{Extratropical\\Cyclone}          & \specialcell{typhoon, storm, farmland,\\ houses, storm coincided}             \\ \hline
Fire                           & \specialcell{fire, hospitals, blaze,\\ water crisis, firefighters}            \\ \hline
Flash Flood                    & \specialcell{flood, floods, flash floods,\\ monitoring stations, muhuri}      \\ \hline
Flood                          & \specialcell{floods, flood, flooding,\\ flood victims, rains}                \\ \hline
Heat Waves                     & \specialcell{heat, temperatures,\\ heat wave, heatstroke,\\ sunstroke}          \\ \hline
Insect Infestation             & \specialcell{locust, food crisis,\\ infestations, millet, harvest}           \\ \hline
Land Slide                     & \specialcell{landslides, houses,\\ hunza river, search, village}          \\ \hline
Mud Slide                      & \specialcell{mudslides, rains, mudslide,\\ torrential rains, houses}         \\ \hline
Other                          & \specialcell{climate, ocean, drought,\\ impacts, warming}                    \\ \hline
\specialcell{Severe Local\\Storm}             & \specialcell{tornado, storm, tornadoes,\\ houses, storms}                   \\ \hline
Snow Avalanche                 & \specialcell{avalanches, avalanche,\\ snow, snowfall,\\ an avalanche}          \\ \hline
Storm Surge                    & \specialcell{king tides, tropical storm,\\ ocean, cyclone season,\\ flooded}    \\ \hline
\specialcell{Technological\\Disaster}         & \specialcell{environmental, toxic waste,\\ pollution, tanker, sludge}      \\ \hline
Tropical Cyclone               & \specialcell{hurricane, cyclone,\\ storm, tropical storm,\\ national hurricane} \\ \hline
Tsunami                        & \specialcell{earthquake, disaster,\\ tsunamis, wave, rains}                   \\ \hline
Volcano                        & \specialcell{eruption, lava, volcanic,\\ crater, eruptions}                   \\ \hline
Wild Fire                      & \specialcell{fires, fire, forest fires,\\ firefighters, burning}  \\ \hline
\end{tabular}
\end{center}
\caption{Words that have the highest mutual information with each disaster type.}
\label{disaster-anchor-words}
\end{table}

\begin{sidewaysfigure}[!th]
\includegraphics[scale = .9, trim = 65 75 0 -175]{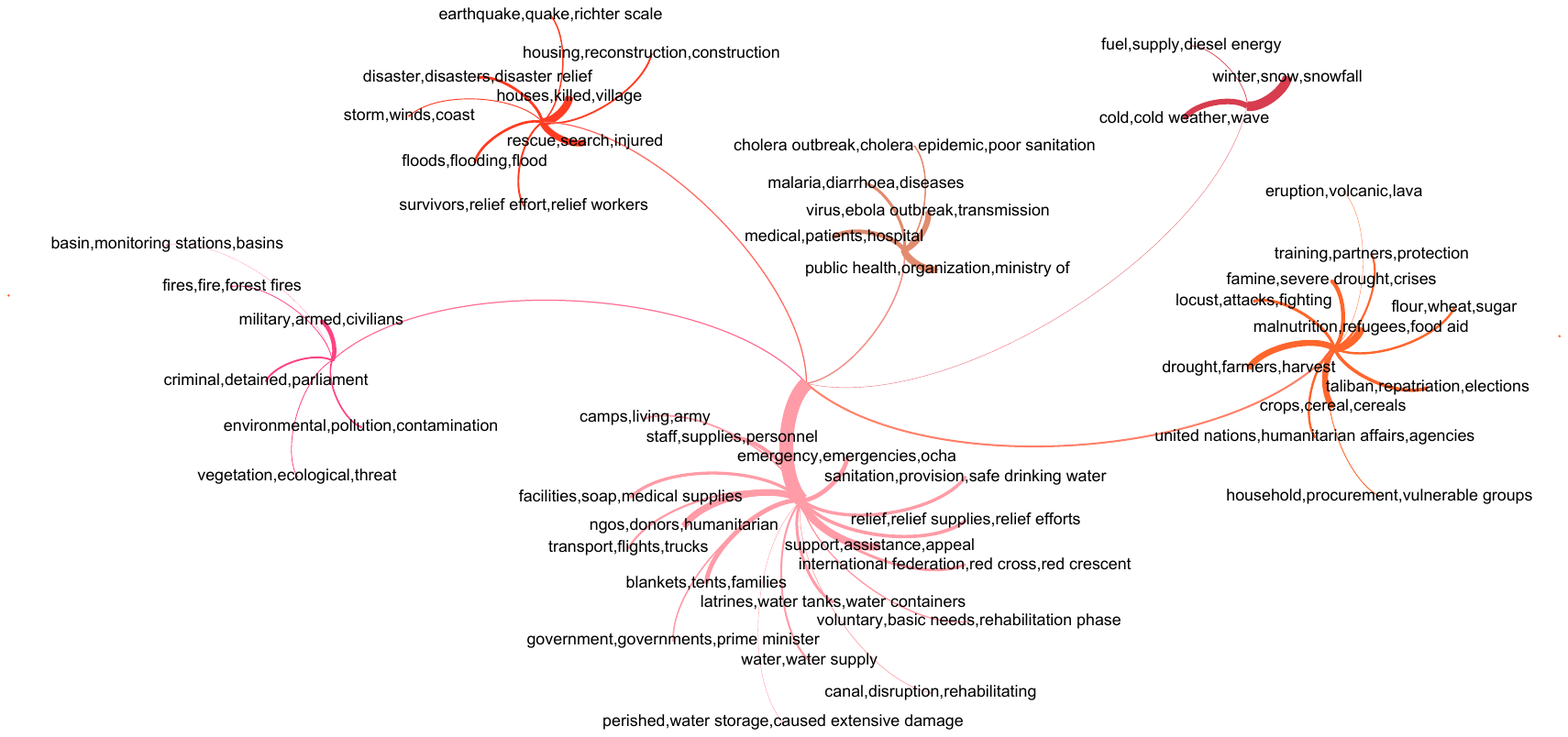}
\caption{Hierarchical CorEx topic model of the disaster relief articles. Edge widths are proportional to the mutual information with the latent representation.}
\end{sidewaysfigure}

\begin{table*}[h]
\normalsize
\begin{center}
\begin{tabular}{|c|l|}
\hline \bf Rank & \bf Topic \\  \hline
1 & drought, farmers, harvest, crop, livestock, planting, grain, maize, rainfall, irrigation \\ \hline
2 & floods, flooding, flood, rains, flooded, landslides, inundated, rivers, submerged, flash floods \\ \hline
3 & \specialcell{eruption, volcanic, lava, crater, eruptions, volcanos, slopes, volcanic activity, evacuated,\\lava flows} \\ \hline
4 & \specialcell{storm, winds, coast, hurricane, weather, tropical storm, national hurricane, coastal, storms,\\ meteorological} \\ \hline
5 & \specialcell{virus, ebola outbreak, transmission, health workers, vaccination, ebola virus, suspected cases,\\ fluids, ebola virus disease, ebola patients} \\ \hline
6 & \specialcell{malnutrition, refugees, food aid, nutrition, feeding, refugees in, hunger, nutritional, refugee,\\ food crisis} \\ \hline
7 & \specialcell{international federation, red cross, red crescent, societies, volunteers, disaster relief\\ emergency, national societies, disaster preparedness, information bulletin, relief operation} \\ \hline
8 & \specialcell{winter, snow, snowfall, temperatures, heavy snow, heating, freezing, warm clothing,\\severe winter, avalanches} \\ \hline
9 & support, assistance, appeal, funds, assist, contributions, fund, cash, contribution, organizations \\ \hline
10 & taliban, repatriation, elections, militia, convoy, ruling, talibans, islamic, convoys, vote \\ \hline
11 & \specialcell{ngos, donors, humanitarian, un agencies, mission, funding, unicef, conduct, humanitarian\\assistance, inter-agency} \\ \hline
12 & fires, fire, forest fires, burning, firefighters, wildfires, blaze, flames, fire fighting, forests \\ \hline
13 & \specialcell{earthquake, quake, richter scale, aftershocks, earthquakes, magnitude earthquake, magnitude,\\ devastating earthquake, an earthquake, earthquake struck} \\ \hline
14 & \specialcell{blankets, tents, families, clothing, utensils, plastic sheeting, clothes, tarpaulins, schools,\\ shelters} \\ \hline
15 & \specialcell{rescue, search, injured, helicopters, death toll, rescue operations, rescue teams, police,\\ rescuers, stranded} \\ \hline
16 & crops, cereal, cereals, millet, food shortages, sorghum, harvests, shortage, ration, rainy \\ \hline
17 & medical, patients, hospital, hospitals, nurses, clinics, clinic, doctor, medical team, beds \\ \hline
18 & \specialcell{water, water supply, drinking water, pumps, drinking, water supplies, potable water, water\\distribution, installed, constructed} \\ \hline
19 & \specialcell{locust, attacks, fighting, infestations, pesticides, opposition, attack, reform, dialogue,\\ governance} \\ \hline
20 & \specialcell{environmental, pollution, contamination, fish, impacts, water quality, polluted, pollutants,\\ chemicals, tanker} \\ \hline
21 & \specialcell{malaria, diarrhoea, diseases, oral, rehydration, salts, contaminated, epidemics, borne diseases,\\ respiratory infections, clean} \\ \hline
22 & \specialcell{emergency, emergencies, ocha, disaster response, coordinating, emergency response,\\ coordinated, coordinators, transportation, rapid assessment} \\\hline
23 & military, armed, civilians, soldiers, aircraft, weapons, rebel, planes, bombs, military personnel \\ \hline
24 & \specialcell{united nations, humanitarian affairs, agencies, agency, governmental, united nations childrens\\ fund, relief coordinator, general assembly, international cooperation, donor community} \\ \hline
25 & transport, flights, trucks, airport, transported, flight, truck, airlift, cargo, route \\ \hline
\end{tabular}
\end{center}
\caption{Topics 1--25 resulting from the best of 10 CorEx topic models run on the disaster relief articles. Topics are ranked by total correlation explained.}
\label{disaster-topics}
\end{table*}

\begin{table*}[h]
\normalsize
\begin{center}
\begin{tabular}{|c|l|}
\hline \bf Rank & \bf Topic \\  \hline
26 & \specialcell{basin, monitoring stations, basins, muhuri, flood forecasting, significant rainfall, moderate\\ rainfall, upstream, light, sludge} \\ \hline
27 & criminal, detained, parliament, protest, crime, protests, protesters, suspects, firing, incident \\  \hline
28 & \specialcell{public health, organization, ministry of, efforts, outbreaks, building, leaders, civil society,\\ minister of, facility} \\ \hline
29 & \specialcell{housing, reconstruction, construction, repair, rebuilding, repairs, temporary housing,\\ corrugated, permanent housing, debris removal} \\ \hline
30 & houses, killed, village, were killed, buildings, swept, debris, roofs, roof, collapse \\ \hline
31 & \specialcell{training, partners, protection, interventions, delivery establishment, violence, benefit,\\ unfpa, pilt} \\ \hline
32 & \specialcell{sanitation, provision, safe, drinking water, latrine, hygiene education, implementing partners,\\ diarrhoeal diseases, rehabilitated, dispaced persons, sanitation services} \\ \hline
33 & flour, wheat, sugar, vegetable, beans, rations, food rations, bread, lentils, needy \\ \hline
34 & camps, living, army, troops, resettlement, relocated, relocation, relocate, flee, settlement \\ \hline
35 & \specialcell{disaster, disasters, disaster relief, cyclone, coordinating council, cyclones, aftermath,\\ devastation, devastated, natural disaster} \\ \hline
36 & \specialcell{relief, relief supplies, relief efforts, relief operations, relief assistance, relief goods, relief\\ materials, relief agencies, donate, providing relief} \\ \hline
37 & \specialcell{household, procurement, vulnerable groups, beneficiary, pipeline, rehabilitate, local ngos,\\ iodised salt, rainfed areas, water harvesting} \\ \hline
38 & \specialcell{staff, supplies, personnel, deployed, staff members, airlifted specialists, flown, logistical\\ support, airlifting} \\ \hline
39 & \specialcell{facilities, soap, medical supplies, clean water, sanitation facilities, emergency medical,\\ international organization, psychosocial, tent, migration iom} \\ \hline
40 & \specialcell{fuel, supply, diesel energy, nitrate, diesel fuel, orphanages, grid, hydroelectric, storage,\\ facilities} \\ \hline
41 & \specialcell{cold, cold weather, wave, warm clothes, extreme temperatures, firewood, severe cold\\ weather, severe cold wave, average temperature} \\ \hline
42 & \specialcell{cholera outbreak, cholera epidemic, poor sanitation, cholera outbreaks, wash, poor hygiene,\\ dirty water, disinfect, hygiene awareness, good hygiene practices} \\ \hline
43 & \specialcell{government, governments, prime minister, administration, national disaster management,\\ corporation, dollars, bilateral donors, disburse, telecom} \\ \hline
44 & \specialcell{famine, severe drought, crises, prolonged drought, devastating, mortality rate, degradation,\\ catastrophic, famine relief, agricultural practices} \\ \hline
45 & \specialcell{vegetation, ecological, threat, mosquitoes, insect, insecticides, lakes, prolonged,\\ habitation, adverse weather} \\ \hline
46 & \specialcell{latrines, water tanks, water containers, affected communities, chlorine tablets, household\\ kits, solid waste, reception centre, local organisations, piped water} \\ \hline
47 & \specialcell{survivors, relief effort, relief workers, survivor, clean drinking water, outlying areas,\\ devastating cyclone, cyclone struck, cyclone survivors, medic} \\ \hline
48 & \specialcell{perished, water storage, caused extensive damage, soil erosion, total loss, sewage systems,\\ salt water, soup, water purifying tablets, electric power} \\ \hline
49 & \specialcell{canal, disruption, rehabilitating, infrastructures, vulnerable areas, uninterrupted, power \\plants, stagnant, inaccessible areas, distress} \\ \hline
50 & \specialcell{voluntary, basic needs, rehabilitation phase, blankets mattresses, raised, freight,\\ humanitarian organizations, government agency, delta region, persons displaced} \\ \hline
\end{tabular}
\end{center}
\caption{Topics 26--50 resulting from the best of 10 CorEx topic models run on the disaster relief articles. Topics are ranked by total correlation explained.}
\label{disaster-topics}
\end{table*}

\clearpage

\begin{sidewaysfigure}[!h]
\includegraphics[scale = .9, trim = 100 65 0 -130]{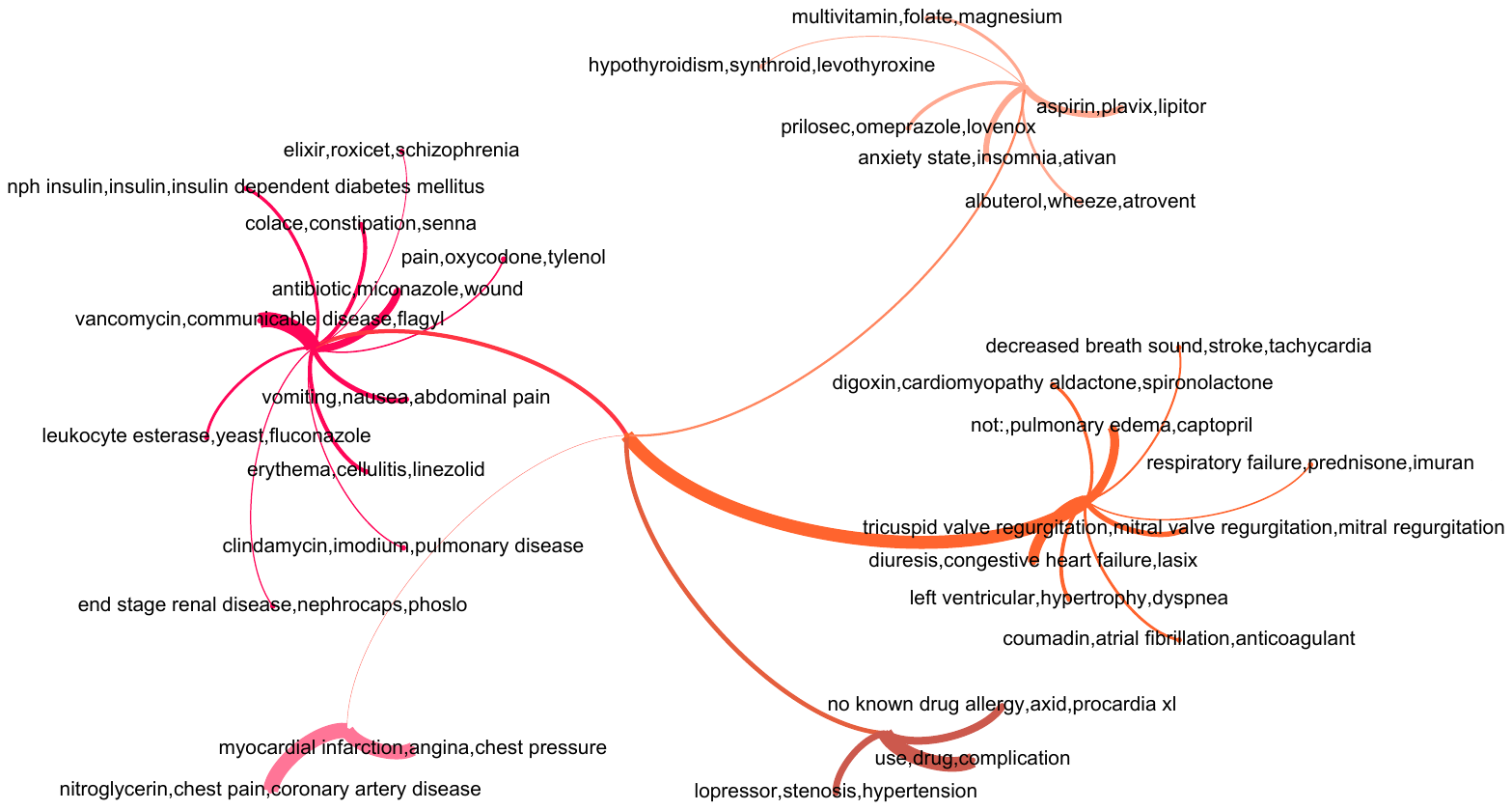}
\caption{Hierarchical CorEx topic model of the clinical health notes. Edge widths are proportional to the mutual information with the latent representation.}
\end{sidewaysfigure}

\clearpage

\begin{table*}[h]
\normalsize
\begin{center}
\begin{tabular}{|c|l|}
\hline \bf Rank & \bf Topic \\  \hline
1 & \specialcell{use, drug, complication, allergy, sodium, infection, furosemide, docusate, shortness of\\ breath, potassium chloride} \\ \hline
2 & \specialcell{vancomycin, communicable disease, flagyl, levofloxacin, diabetes, renal failure, sepsis,\\ ceftazidime, nutrition, gentamicin} \\ \hline
3 & \specialcell{aspirin, plavix, lipitor, toprol xl, lantus, hydroxymethylglutaryl coa reductase inhibitors,\\ atorvastatin, nexium, novolog, disease} \\ \hline
4 & \specialcell{diuresis, congestive heart failure, lasix, edema, orthopnea, crackle, heart failure, dyspnea on\\ exertion, oxygen, torsemide} \\ \hline
5 & \specialcell{albuterol, wheeze, atrovent, chronic obstructive pulmonary disease, asthma, flovent,\\ ipratropium, fluticasone, advair, combivent} \\ \hline
6 & \specialcell{end stage renal disease, nephrocaps, phoslo, calcitriol, cellcept, kidney transplant,\\ arteriovenous fistula, acetate, cyclosporine, neoral} \\ \hline
7 & \specialcell{nitroglycerin, chest pain, coronary artery disease, hypokinesia, st depression, lesion,\\ unstable angina, akinesia, st elevation, diaphoresis} \\ \hline
8 & \specialcell{respiratory failure, prednisone, imuran, immunosuppression, necrosis, cyclosporin, sick,\\ magnesium oxide, tachypnea, arteriovenous malformation} \\ \hline
9 & \specialcell{elixir, roxicet, schizophrenia, risperdal, zofran, crushed, valproic acid, promethazine,\\ phenergan, prochlorperazine} \\ \hline
10 & \specialcell{leukocyte esterase, yeast, fluconazole, urosepsis, dysphagia, oxycontin, lidoderm,\\ chemotherapy, adriamycin, medical problems} \\ \hline
11 & \specialcell{colace, constipation, senna, lactulose, dulcolax, milk of magnesia, sennoside, dilaudid,\\ protonix, reglan} \\ \hline
12 & \specialcell{vomiting, nausea, abdominal pain, diarrhea, fever, dehydration, chill, clostridium difficile,\\ intravenous fluid, compazine} \\ \hline
13 & \specialcell{coumadin, atrial fibrillation, anticoagulant, warfarin, k vitamin, amiodarone, atrial flutter,\\ flutter, deep venous thrombosis, allopurinol} \\ \hline
14 & \specialcell{digoxin, cardiomyopathy, aldactone, spironolactone, carvedilol, dobutamine, alcohol,\\ idiopathic cardiomyopathy, ventricular rate, addiction} \\ \hline
15 & \specialcell{clindamycin, imodium, pulmonary disease, erythromycin, defervesced, sweating, carafate,\\ quinidine, cytomegalovirus, cepacol} \\ \hline
16 & \specialcell{lopressor, stenosis, hypertension, heparin, hypercholesterolemia, aortic valve insufficiency,\\ mitral valve insufficiency, aortic valve stenosis, sinus rhythm, peripheral vascular disease} \\ \hline
17 & \specialcell{antibiotic, miconazole, wound, nitrate, morbid obese, fentanyl, sleep apnea, obesity, abscess,\\ ampicillin} \\ \hline
18 & \specialcell{erythema, cellulitis, linezolid, swelling, erythematous, osteomyelitis, ancef, keflex,\\ dicloxacillin, bacitracin} \\ \hline
19 & \specialcell{anxiety state, insomnia, ativan, neurontin, depression, lorazepam, gabapentin, trazodone,\\ fluoxetine, headache} \\ \hline
20 & \specialcell{multivitamin, folate, magnesium, folic acid, mvi, maalox, thiamine, vitamin c, gluconate,\\ dyspepsia} \\ \hline
\end{tabular}
\end{center}
\caption{Topics 1--20 resulting from the best of 10 CorEx topic models run on the clinical health notes. Topics are ranked by total correlation explained.}
\label{disaster-topics}
\end{table*}

\begin{table*}[t!]
\normalsize
\begin{center}
\begin{tabular}{|c|l|}
\hline \bf Rank & \bf Topic \\  \hline
21 & \specialcell{decreased breath sound, stroke, tachycardia, seizure disorder, lymphocyte, atelectasis,\\ polymorphonuclear leukocytes, ecchymosis, seizure, cefotaxime} \\ \hline
22 & \specialcell{not: , pulmonary edema, captopril, pleural effusion, rales, beta blocker, fatigue, dead,\\ q wave, dysfunction} \\ \hline
23 & \specialcell{hypothyroidism, synthroid, levothyroxine, levoxyl, diovan, valsartan, angioedema,\\ bestrophinopathy, atherosclerosis, ursodiol} \\ \hline
24 & \specialcell{nph insulin, insulin, insulin dependent diabetes mellitus, anemia, humulin insulin,\\ retinopathy, hyperglycemia, humulin, gastrointestinal bleeding, nephropathy} \\ \hline
25 & \specialcell{tricuspid valve regurgitation, mitral valve regurgitation, mitral regurgitation, left atrial\\ enlargement, zaroxolyn, ectopy, right atrial enlargement, metolazone, deficit, regurgitant} \\ \hline
26 & \specialcell{prilosec, omeprazole, lovenox, pulmonary embolism, enoxaparin, xalatan, oxybutynin,\\ helicopter pylori, flonase, ramipril} \\ \hline
27 & \specialcell{pain, oxycodone, tylenol, percocet, ibuprofen, morphine, osteoarthritis, hernia, motrin,\\ bleeding} \\ \hline
28 & \specialcell{left ventricular hypertrophy, dyspnea, living alone, smokes, syndrome, hives, palpitation,\\ elderly, left axis deviation, usual state of health} \\ \hline
29 & \specialcell{myocardial infarction, angina, chest pressure, patent ductus arteriosus, atenolol, micronase,\\ adenosine, non-insulin dependent diabetes mellitus, ecotrin, ~caltrate} \\ \hline
30 & \specialcell{no known drug allergy, axid, procardia xl, vasotec, obese, mevacor, tissue plasminogen\\activator, middle-aged, nifedipine, procardia} \\ \hline
\end{tabular}
\end{center}
\caption{Topics 21--30 resulting from the best of 10 CorEx topic models run on the clinical health notes. Topics are ranked by total correlation explained.}
\label{disaster-topics}
\end{table*}

\end{document}